\documentclass{bmvc2k}

\usepackage{tipa}
\usepackage{amssymb,amsmath,bm}
\usepackage{todo}
\usepackage{placeins}

\title{Visual gesture variability between talkers \\ in continuous visual speech}

\addauthor{Helen L Bear}{helen@uel.ac.uk}{1}

\addinstitution{
University of East London\\
 London E16 2RD UK
}

\runninghead{Bear}{Visual Gesture Variability}

\begin{document}

\maketitle

\begin{abstract}
Recent adoption of deep learning methods to the field of machine lipreading research gives us two options to pursue to improve system performance. Either, we develop end-to-end systems holistically or, we experiment to further our understanding of the visual speech signal. The latter option is more difficult but this knowledge would enable researchers to both improve systems and apply the new knowledge to other domains such as speech therapy. 

One challenge in lipreading systems is the correct labeling of the classifiers. These labels map an estimated function between visemes on the lips and the phonemes uttered. Here we ask if such maps are speaker-dependent? Prior work investigated isolated word recognition from speaker-dependent (SD) visemes, we extend this to continuous speech. Benchmarked against SD results, and the isolated words performance, we test with RMAV dataset speakers and observe that with continuous speech, the trajectory between visemes has a greater negative effect on the speaker differentiation. 
\end{abstract}

\section{Introduction}
In machine lip-reading we have a dilemma: evidence suggests we can identify individuals from their unique visual speech information (\cite{607030,1703580}) yet with conventional systems we pursue the ideal scenario of being able to lip-read any speaker. Recent work with deep learning classifiers and very large datasets promise to achieve this in end-to-end systems but by taking this road, we risk failing to understand the visual speech signal. Whilst there are behavoural studies on human perception of visual speech (human speech readers), we are yet to fully utilise computers fully to undertake these investigations. 

In machine lip-reading, one attempts to interpret words spoken from the visual representation of sounds as they're uttered. This means there are two levels of `translation' within the process, visual gestures (known as visemes) into phonemes, and phonemes into words. Previous literature reports lip-reading performance based on different units in these three stages which makes comparing performance difficult. Some report word error rate \cite{lucey2009visual,4064511}, others viseme error rate \cite{bear2014resolution}, or others accuracy of these two units. However, without a known function, or map, to determine which phoneme a viseme translates to (some phonemes are visually indistinguishable and there are many suggested mappings in literature), and the same words can be pronounced with alternative phonemes, there is an argument for a speaker-dependent function for mapping between visual gestures and phonemes \cite{bear2016decoding}. This idea was first presented in \cite{visualvowelpercept} and the latest algorithm for deriving speaker-dependent maps from phoneme confusions was presented in \cite{bear2014phoneme}. But we need to understand, how speaker-dependent should these maps be? Whilst a speaker-generic phoneme-to-viseme (P2V) map is desirable as the concept of a viseme alphabet for every speaker is daunting, we know that adapting a trained classifier from one speaker to another similar speaker, is more data-efficient than learning a model for all speakers \cite{cox2008challenge,kleinschmidt2015robust} which in turn achieves less accuracy. Therefore, we try to determine, just how different are speaker-dependent visemes when lip-reading continuous speech? The only prior attempt at answering this question used an isolated word dataset \cite{bear2015speakerindep}. Here we use a similar methodology to test the effects of the P2V maps and suggest a simple metric to compare the maps. 

\section{Method Overview}
We use the phoneme clustering approach in \cite{bear2014phoneme} to produce a series of speaker-dependent P2V maps. The intention behind the design of this viseme derivation algorithm, is to utilise the unique gestures of each speaker to improve lip reading over previous viseme sets \cite{taylor2012dynamic} and phoneme labelled classifiers \cite{bear2014phoneme}. Our series of maps is made up of the following: 
\begin{enumerate} 
\item a multi-speaker P2V map using {\em all} speakers' phoneme confusions; \vspace{-4pt}
\item a speaker-independent P2V map for each speaker using confusions of all {\em other} speakers in the data; \vspace{-4pt}
\item a speaker-dependent P2V map for each speaker.
\end{enumerate} 
So we have 25 P2V maps (one multi-speaker map, and 12 speaker maps per types two and three). Maps are constructed using separate training and test data over ten fold cross-validation \cite{efron1983leisurely}. All 25 P2V maps are in the appendices of \cite{thesis}. 

We use the HTK toolkit \cite{young2006htk} and build HMM classifiers with the visemes in each P2V map. The HMMs are flat-started with \texttt{HCompV}, re-estimated $11$ times over (\texttt{HERest}) with forced alignment after the seventh re-estimate. Then we classify using \texttt{HVite} and output results with \texttt{HResults}. We follow the design decisions of \cite{Matthews_Baker_2004}, with three state HMMs with an associated five-component Gaussian mixture per state \cite{bear2015speakerindep}. A bigram word network is built with \texttt{HBuild} and \texttt{HLStats}, and classification is measured as word Correctness (Equation~\ref{eq:correctness}) \cite{young2006htk}.  We report word correctness, $C_w$, rather than viseme correctness. This is because lip-reading is the understanding of speech, i.e. the interpretation of words rather than the matching of a single viseme. This also normalises for training sample bias across the viseme sets. The number of visemes varies per speaker so training sample volumes are not consistent, but the number of words spoken is. The BEEP pronunciation dictionary used throughout these experiments is in British English \cite{beep}.
\begin{equation} 
C_w = \displaystyle \frac{N-D-S}{N}\quad 
\label{eq:correctness}
\end{equation} 
In Equation~\ref{eq:correctness} $C_w$ is the word correctness, $N$ is the total number of words, $D$ is the number of deletion errors, and $S$ are substitution errors. These are counted when comparing our classifier recognition outputs with the ground truth. 

\subsection{RMAV Data}
Of the continuous speech datasets available, most still use speaker-dependent tests \cite{lan2012insights, Hazen1027972, Wong20111503, cappelletta2012phoneme}. Some are single speaker-dependent, others multi-speaker dependent but the point is that the test speaker is included in the training data (not the same samples). Active Appearance Models (AAMs) have been described fully in previous literature \cite{Matthews_Baker_2004} so we do not repeat this here but we track all 12 speakers and extract AAM features. 

The RMAV dataset is a corpus of $12$ speakers, seven male and five female, each reciting $200$ sentences selected from the Resource Management Corpus [3]. The database has a vocabulary size of approximately $1000$ words, and was recorded in full-frontal view in HD. Head motion is restrained and lighting constant. 

\section{Experiment Design}
Each test in this experiment is designated as: 
\begin{equation} 
M_n(p,q)\quad 
\label{eq2} 
\end{equation} 
This means P2V map $M_n$ from speaker $n$, is trained using visual speech data from speaker $p$ and tested using speaker $q$. E.g. $M_1(2,3)$ designates testing a P2V map constructed from Speaker 1, using training data from Speaker 2, and testing on Speaker 3. 

\subsection{Multi-speaker (MS) and Speaker-Independent (SI) tests}
Accuracte speaker-dependent lip-reading exists \cite{cappelletta2012phoneme,thangthai2015improving}. However with independent speakers between training and test sets, accuracy significantly falls \cite{cox2008challenge}. Our first tests use a P2V map based on the phoneme confusions of all speakers. Therefore, the MS map is tested as: $M_{[all]}(1,1)$, $M_{[all]}(2,2)$, $M_{[all]}(3,3)$, $M_{[all]}(4,4)$, $M_{[all]}(5,5)$, $M_{[all]}(6,6)$, $M_{[all]}(7,7)$, $M_{[all]}(8,8)$, $M_{[all]}(9,9)$, $M_{[all]}(10,10)$, $M_{[all]}(11,11)$, and $M_{[all]}(12,12)$. 
Our SI tests use 12 maps derived using all speakers confusions bar the test speaker. This time we substitute the symbol `!$n$' in place of a list of speaker ID numbers, meaning `not including speaker $n$'. The tests for these maps are as follows $M_{!1}(1,1)$, $M_{!2}(2,2)$ and so on. 

\subsection{Different Speaker-Dependent maps \& Data (DSD\&D) tests}
This set of tests use the P2V maps and training data of the non-test speaker. So for Speaker 1 we test $M_2(2,1)$, $M_3(3,1)$, up to $M_12(12,1)$, and for Speaker 2 we test $M_1(1,2)$, $M_3(3,2)$ and so on for all speakers. Table~\ref{tab:sid_l} shows Speaker 1 tests.

\begin{table}[!h] 
\centering 
\caption{Different speaker-dependent maps and Data (DS\&D) experiments for Speaker one.} 
 \resizebox{\columnwidth}{!}{%
 \begin{tabular}{| l | l | l | l |}
\hline 
Mapping ($M_n$) & Training data ($p$) & Test speaker ($q$) & $M_n(p,q)$ \\ 
\hline
Sp2 & Sp2 & Sp1 & $M_2(2,1)$ \\ 
Sp... & Sp... & Sp1 & $M_...(...,1)$ \\ 
Sp12 & Sp12 & Sp1 & $M_{12}(12,1)$ \\ 
\hline 
\end{tabular} %
}
\label{tab:sid_l} 
\end{table} 

\subsection{Different Speaker-Dependent maps (DSD) tests}
These are the same speaker-dependent P2V maps as tested in Table~\ref{tab:sid_l} but in these tests we trained and test on the same speaker (maintaining independent samples between train and test folds). In Table~\ref{tab:si_l} we show the DSD tests for Speaker 1 as an example. 

\begin{table}[!ht] 
\centering 
\caption{Different Speaker-Dependent maps (DSD) for Speaker one.} 
 \resizebox{\columnwidth}{!}{%
 \begin{tabular}{| l | l | l | l |} 
\hline 
Mapping ($M_n$) & Training data ($p$) & Test speaker ($q$) & $M_n(p,q)$ \\ 
\hline 
Sp2 & Sp1 & Sp1& $M_2(1,1)$ \\ 
Sp... & Sp1 & Sp1 & $M_...(1,1)$ \\ 
Sp12 & Sp1 & Sp1 & $M_{12}(1,1)$ \\ 
\hline 
\end{tabular} %
}
\label{tab:si_l} 
\end{table} 

\subsection{Speaker dependent lip-reading benchmark (SSD)}
Our baseline performance are speaker-dependent P2V results where $n=p=q$. They are SSD because each map,  training and test data are all from the same speaker. This mimics previous work in conventional systems due to small AV datasets. These maps are: $M_1(1,1)$, $(M_2(2,2)$, $(M_3(3,3)$, up to $M_12,(12,12)$.

\section{Analysis of Results}
All results are measured in $C_w\%$ with error bars showing one standard error (s.e) over $10$ folds. 
\subsection{Multi-speaker and speaker-independent maps}
\begin{figure}[!h] 
\centering 
\includegraphics[width=0.95\linewidth]{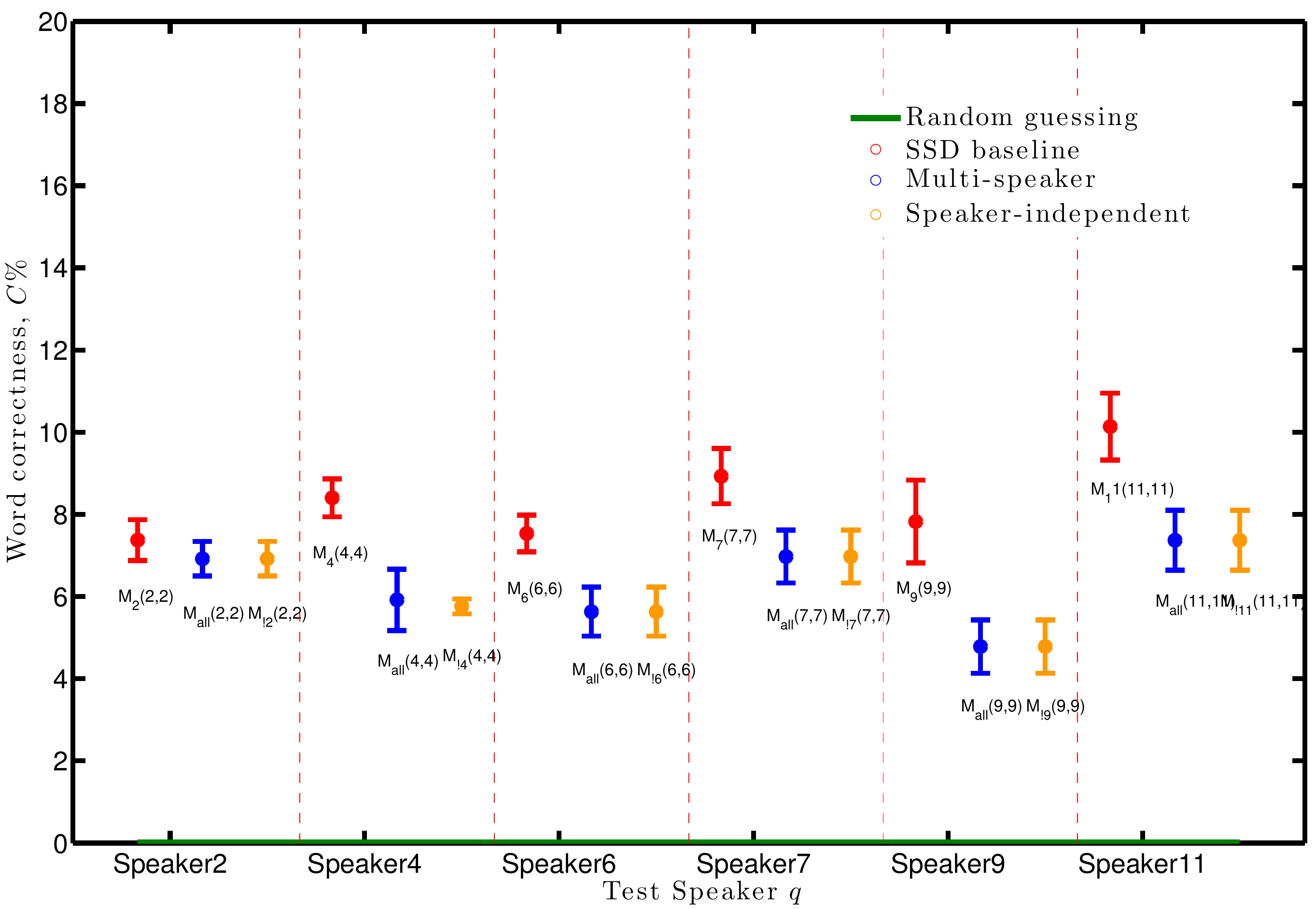} 
\caption{Word classification correctness, $C\pm1s.e$, of the MS and SI tests. Baseline is SSD maps (red)} 
\label{fig:accuracy1} 
\end{figure} 

In Figures~\ref{fig:accuracy1} we have plotted the MS \& SI experiments with representative example speakers (all results are plotted in supplementary materials). The SD benchmark scores are in red. All speakers bar Speaker 2 are significantly negatively affected by using generalised multi-speaker visemes, including test speakers phoneme confusions in the MS P2V doesn't improve $C_w$. This quantifies lip-reading dependency on speaker identity as dependent on which two speakers are being compared. Our one exception, Speaker 2, (Figure~\ref{fig:accuracy1}) shows only an insignificant decrease in $C_w$ with MS and SI visemes. So one future improvement possibility is deriving multi-speaker visemes based upon sets of visually similar speakers. The challenge remains in knowing which speakers should be grouped together. This is consistent with the neuroscience learning adage how the human brain learns ``\textit{we recognise the familiar, generalise to the similar, and adapt to the novel}'' \cite{kleinschmidt2015robust}. Also, in \cite{bear2015speakerindep} Figure 4 this test on isolated words were $50/50$ if MS or SI maps significantly reduced lipreading. On such a small dataset we can not be conclusive, but it is reassuring that our continuous speech results are similar.

\FloatBarrier
\subsection{Different Speaker-Dependent results}
\begin{figure}[!ht] 
\centering 
\includegraphics[width=0.95\linewidth]{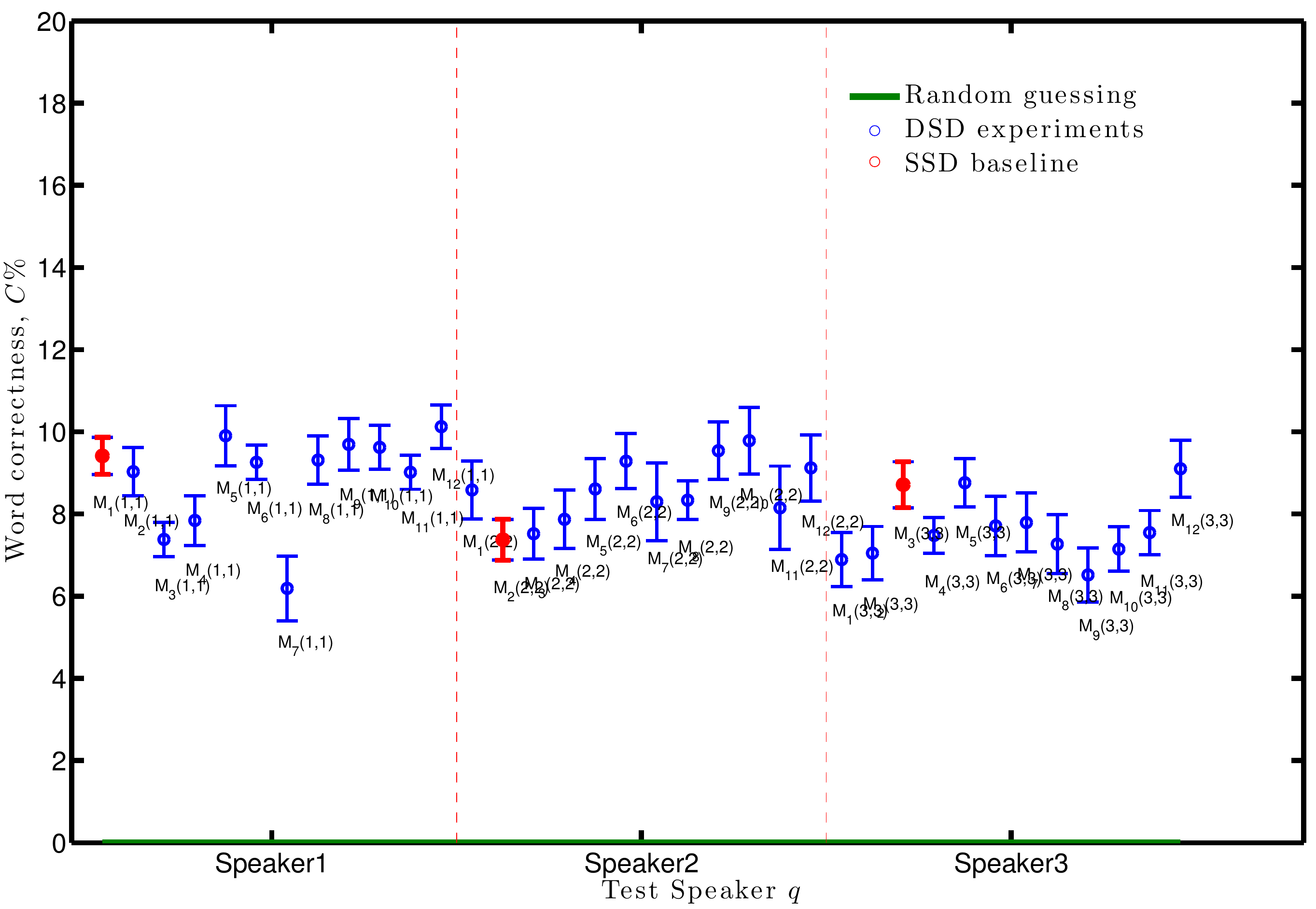} 
\caption{Word classification correctness, $C\pm1s.e$, of the DSD tests for speakers 1, 2 and 3. Baseline is SSD maps (red)} 
\label{fig:correctness1} 
\end{figure}

\begin{figure}[!ht] 
\centering 
\includegraphics[width=0.95\linewidth]{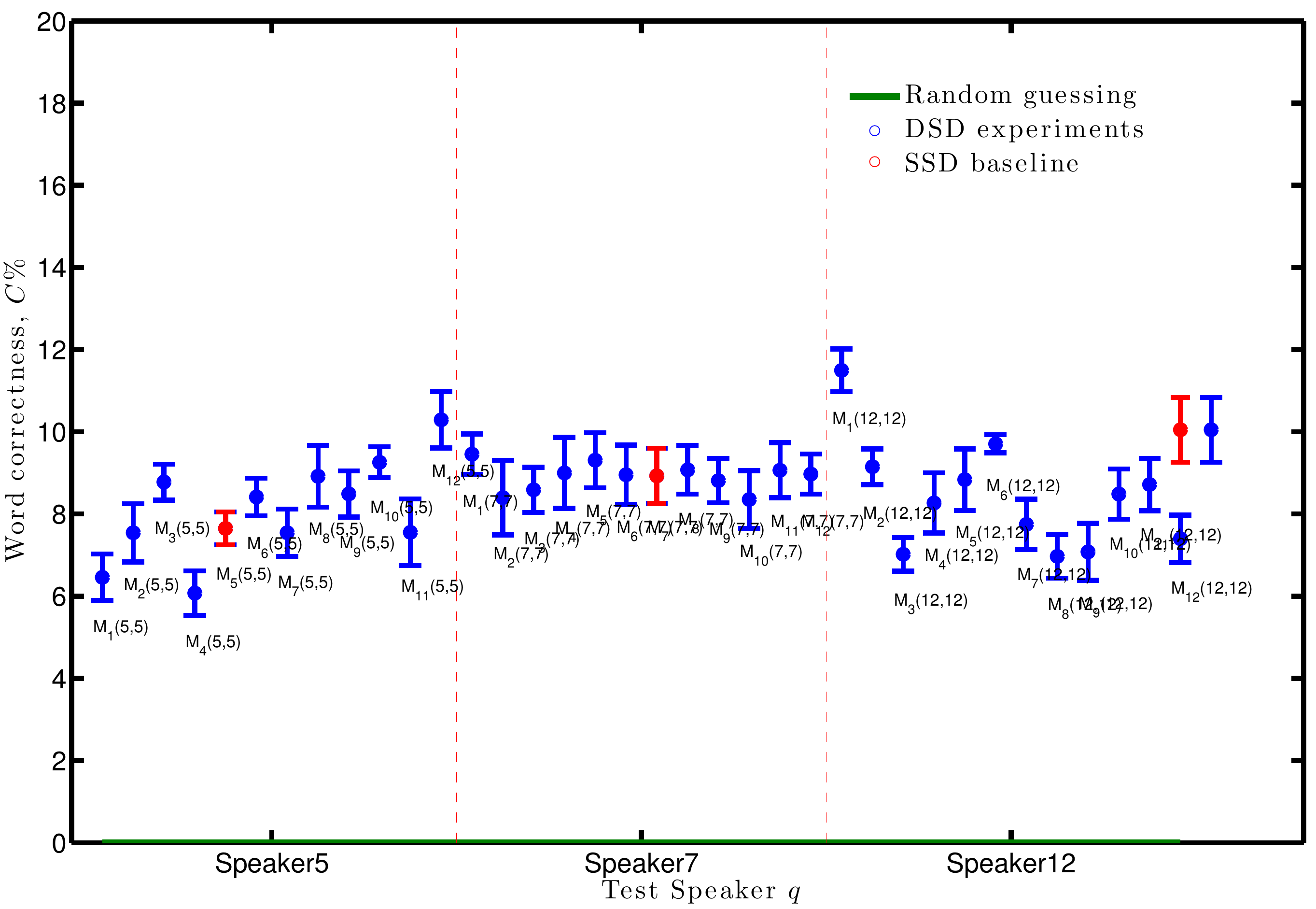} 
\caption{Word classification correctness, $C\pm1s.e$, of the DSD tests for speakers 5, 7 and 12. Baseline is SSD maps (red)} 
\label{fig:correctness4} 
\end{figure}

In Figures~\ref{fig:correctness1} and~\ref{fig:correctness4} we have plotted our results in $C_w$, on the $y$-axis and our SD benchmark. Here, the HMM is trained on the test speaker so we see the effects of the unit selection. In Figure~\ref{fig:correctness1} the three maps $M_3, M_7$ and $M_{12}$ all significantly reduce $C_w$ for Speaker 1. In contrast, for Speaker 2 there are no significantly reducing maps but maps $1$, $4$, $5$, $6$, $9$, and $11$ all significantly improve the classification of Speaker 2. This suggests its not just the speakers identity which is important for good classification but how it is used. Some individuals may simply be easier to lip-read (for reasons as yet unknown) or there are similarities between certain speakers which when learned from Speaker $A $ are adaptable to lip-read visually-similar speakers.
 
In Figure~\ref{fig:correctness4}, Speaker 7 is particularly robust to visual unit selection for the classifier labels. Conversely Speaker's 5 and 12 are significantly affected by the visemes. This variability has not been previously considered, as some speakers may be dependent on good visual classifiers, but others not so much. Again, the number of visual classifiers varies with speaker identity. 

\FloatBarrier
We rank each speaker viseme set by weighting the effect of the DSD tests in Table~\ref{tab:weighting_lilir}. If a map increases on SSD $C_w$ within $1s.e$, this scores $+1$ or outside $1s.e$ scores $+2$. If a map decreases $C_w$, these weights are negative. 
\begin{table} [!h]
\centering 
\caption{Comparison scores measuring the effect of using speaker-dependent maps for \emph{other} speakers lip-reading.} 
 \resizebox{\columnwidth}{!}{%
\begin{tabular}{lrrrrrrrrrrrr} 
\hline
& $M_1$ & $M_2$ & $M_3$ & $M_4$ & $M_5$ & $M_6$ & $M_7$ & $M_8$ & $M_9$ & $M_{10}$ & $M_{11}$ & $M_{12}$ \\ 
\hline
Sp01 & $0$ 		& $-1$		& $-2$		& $-2$		& $+1$		& $-1$		& $-1$		& $-1$		& $+1$		& $+1$		& $-1$		& $+1$ \\ 
Sp02 & $+2$ 		& $0$		& $+1$		& $+1$		& $+2$		& $+2$		& $+1$		& $+1$		& $+2$		& $+2$		& $+1$		& $+2$ \\ 
Sp03 & $-2$ 		& $-2$		& $0$		& $-2$		& $+1$		& $-1$		& $-1$		& $-2$		& $-2$		& $-2$		& $-2$		& $+1$ \\ 
Sp04 & $-2$ 		& $-1$		& $-1$		& $0$		& $+1$		& $+1$		& $-2$		& $-2$		& $+1$		& $-1$		& $-2$		& $+1$ \\ 
Sp05 & $-2$ 		& $-1$		& $+2$		& $-2$		& $0$		& $+1$		& $-1$		& $+2$		& $+1$		& $+2$		& $-1$		& $+2$ \\ 
Sp06 & $-1$ 		& $-1$		& $-1$		& $+1$		& $+2$		& $0$		& $+2$		& $-1$		& $-1$		& $+1$		& $+1$		& $+2$ \\ 
Sp07 & $+1$ 		& $-1$		& $-1$		& $+1$		& $+1$		& $+1$		& $0$		& $+1$		& $-1$		& $-1$		& $+1$		& $+1$ \\ 
Sp08 & $-1$ 		& $-1$		& $+1$		& $-1$		& $-1$		& $-2$		& $-2$		& $0$		& $+1$		& $+2$		& $+1$		& $+1$ \\ 
Sp09 & $-2$ 		& $-2$		& $-1$		& $-2$		& $-1$		& $-1$		& $-1$		& $-2$		& $0$		& $-1$		& $-2$		& $+1$ \\ 
Sp10 & $-2$ 		& $-2$		& $-1$		& $-1$		& $-1$		& $-2$		& $-2$		& $-2$		& $-2$		& $0$		& $-2$		& $-2$ \\ 
Sp11 & $-1$ 		& $+1$		& $-1$		& $+1$		& $+1$		& $-1$		& $+1$		& $-1$		& $-1$		& $+2$		& $0$		& $+2$ \\ 
Sp12 & $-1$ 		& $-2$		& $-2$		& $-1$		& $-1$		& $-2$		& $-2$		& $-2$		& $-2$		& $-1$		& $-2$		& $0$ \\ 
\hline
Total	 & $-9$		& $-11$		& $-6$		& $-7$ 		& $\textbf{+3}$ 		& $-5$ 		& $-8$ 		& $-9$ 		& $-3$ 		& $-4$ 		& $-8$ 		& $\textbf{+12}$ \\ 
\hline
\end{tabular} %
}
\label{tab:weighting_lilir} 
\end{table} 
The key observation in Table~\ref{tab:weighting_lilir} is Speaker 12. $M_{12}$ is one of two ($M_{12}$ and $M_5$) which make an overall improvement in classifying other speakers with positive values in the total row and crucially, $M_{12}$ only has one speaker (Speaker 10) for whom the visemes in $M_{12}$ does not make an improvement in classification. The one other speaker P2V map which improves over other speakers is $M_5$. All others show a negative effect, this reinforces the assertion visual speech is unique to speakers but we now have evidence of exceptions. In future a way of measuring the similarity between the viseme maps of each individual speaker, this would help adaptation from one speaker to another with less training data. 

\subsection{Different Speaker-Dependent map \& Data results}
\begin{figure}[!ht] 
\centering 
\includegraphics[width=0.95\linewidth]{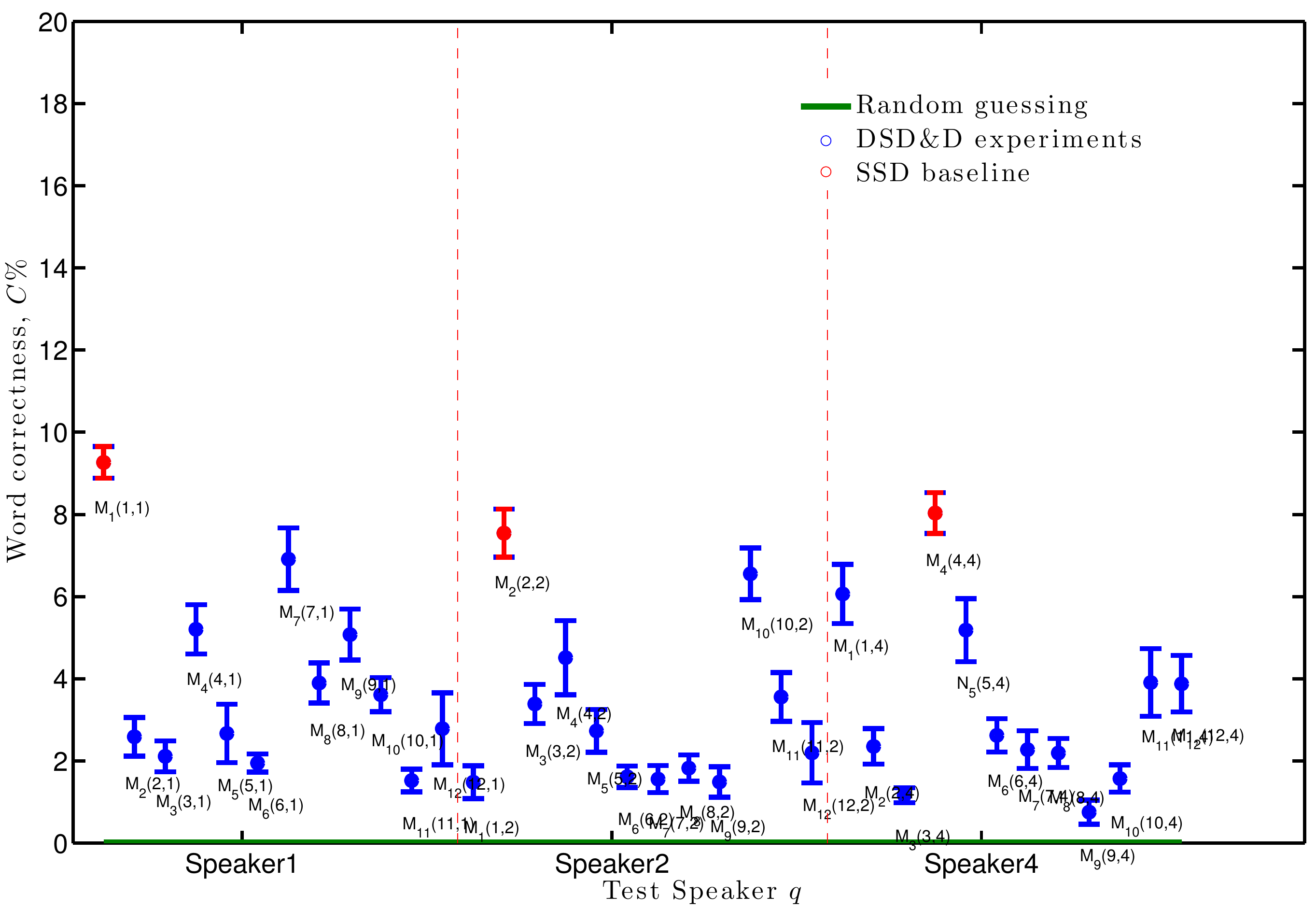} 
\caption{Word classification correctness, $C\pm1s.e$, of the DSD\&D tests for speakers 1, 2 and 4. Baseline is SSD maps (red)}
\label{fig:indep_CorrL1} 
\end{figure}

\begin{figure}[!ht] 
\centering 
\includegraphics[width=0.95\linewidth]{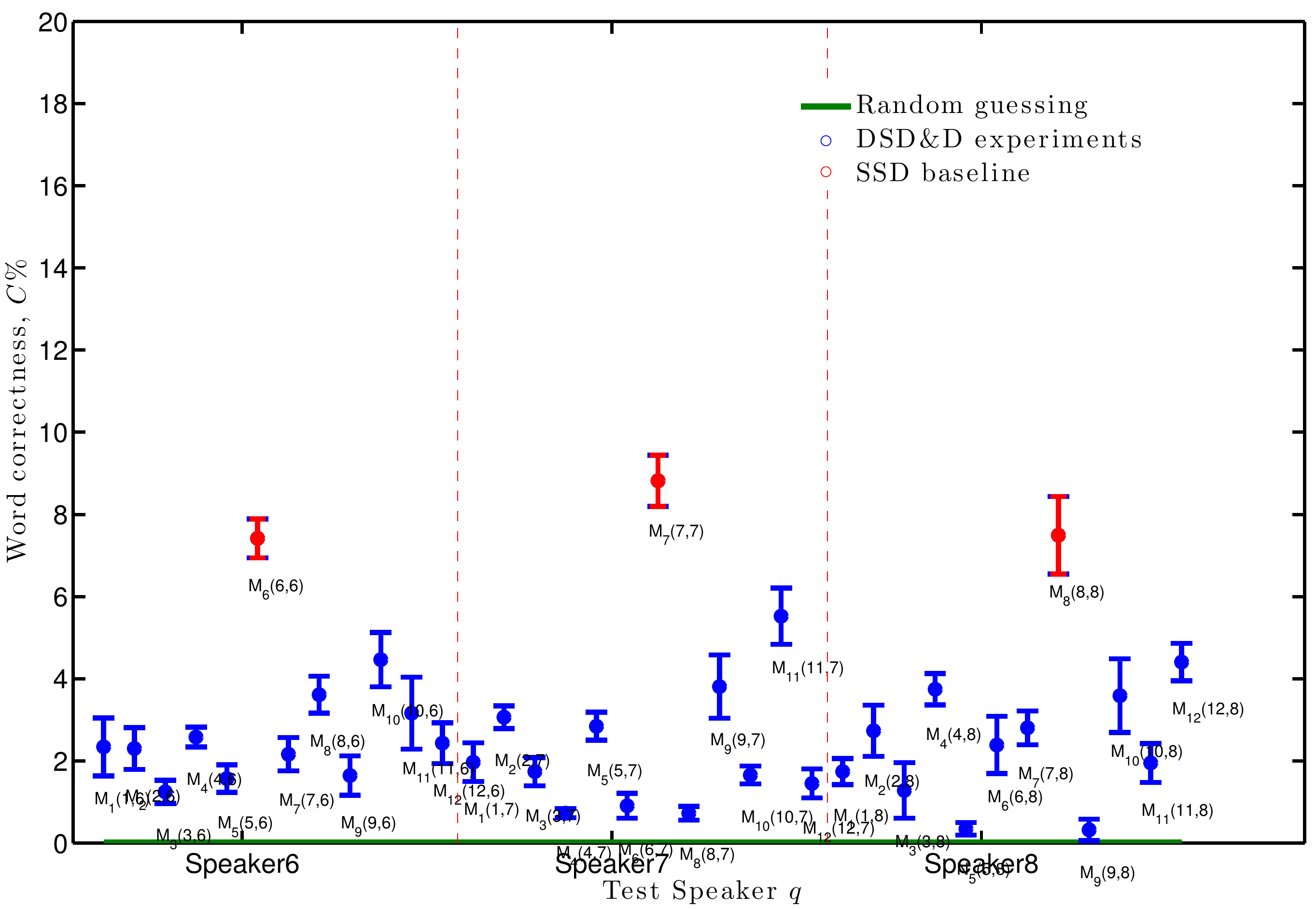} 
\caption{Word classification correctness, $C\pm1s.e$, of the DSD\&D tests for speakers 6,7, and 8. Baseline is SSD maps (red)} 
\label{fig:indep_CorrL4} 
\end{figure}

Figures~\ref{fig:indep_CorrL1} and ~\ref{fig:indep_CorrL4} show the $C_w$ achieved with the labelled DS\&D tests. It is reassuring to see some speakers significantly deteriorate the classification rates when the speaker used to train the classifier is not the same as the test speaker as this is consistent with our MS and SI tests. For example, on the leftmost side of Figure~\ref{fig:indep_CorrL1}, the test speaker is Speaker 1. The speaker-dependent maps for all 12 speakers have been used to build sets of classifiers. But when tested on Speaker 1, only maps and models for speakers 3, 7 and 12 show a significant reduction in word correctness. All eight other speakers are within one standard error. Figure~\ref{fig:indep_CorrL4} we see a similar trend with Speaker 4 showing the most variation of these three speakers. To lip-read Speaker 4 we actually see a significant improvement by using the map and model of Speaker 6 and less significant improvements by speakers 3, 5 and 11. However, whilst these are all signs towards speaker-independent lip-reading, the most common trend is, there is a lot of overlap between our continuous speech speakers and this natural variation is attributed to the speaker identity and or linguistics restrictions. 

If we compare these figures to the isolated words results in \cite{bear2015speakerindep}, either the extra data in this larger data set or the longer sentences in continuous speech have made a difference. Whilst the lack of training on a test speaker is still significantly less accurate than speaker dependent tests, with continuous speech, it is significantly better than the difference experienced in isolated words. Table~\ref{tab:diffs} lists the differences for all speakers on both datasets and the mean difference between isolated words and continuous speech is $C_w\%$. Furthermore, with isolated words, the performance attained by speaker-independent tests was shown in cases to be worse than guessing. Whilst our poorest P2V maps might be low, they are all significantly better than guessing regardless of the test speakers.

\begin{table}
\centering
\caption{}
 \resizebox{\columnwidth}{!}{%
\begin{tabular}{l r r r r r r r r r r r r}
\hline
Dataset	 & Sp1 	& Sp2 	& Sp3 	& Sp4 	& Sp5  	& Sp6 	& Sp7 	& Sp8 	& Sp9 	& Sp10 	& Sp11 	& Sp12 \\
\hline 
AVL2  	& 14.06  	& 11.87	& 42.08	& 32.75 	&	-	& 	-	&	-	&	-	&	-	&	-	&	-	& - \\
RMAV 	&  5.78	&  4.74	& 6.49   	& 5.13   	& 5.57    	& 4.92   	& 6.60   	& 5.19	& 5.64    	& 7.03	&  7.49    	& 8.04 \\
\hline
\end{tabular} %
}
\label{tab:diffs}
\end{table}

\FloatBarrier

\section{Conclusions}
In our MS and SI maps, there are $\pm2$ visemes per set whereas the SSD maps have a range of six. So we conclude, there is high risk of over-generalising a MS/SI P2V map. We suggest it is not only the speaker-dependency that varies but also the contribution of each viseme within the set which affects the word classification performance, an idea also shown in~\cite{bear2014obs}. These results present, that a set of certain MS visemes with some SD visemes, may improve speaker-independent lip-reading.  We have shown exceptions where the P2V map selection is significant and where classifiers trained on non-test speakers has not been detrimental. This gives hope that with visually similar speakers, speaker independent lip-reading is possible. 

Furthermore, with continuous speech, we have shown that speaker dependent phoneme-to-viseme maps significantly improve lipreading over isolated words. We attribute this to the co-articulation effects on phoneme recognition confusions which in turn influences the speaker-dependent maps with linguistic or context information. This is supported by evidence from conventional lipreading systems which show the strength of language models in lipreading accuracy. 

Finally, we have more evidence that speaker independence, even with unique trajectories between visemes for individual speakers, is likely to be achievable. What we need now is more understanding of the influence of language on visual gestures.  

%
%

\bibliography{mybib}

\begin{thebibliography}{24}
\providecommand{\natexlab}[1]{#1}
\providecommand{\url}[1]{\texttt{#1}}
\expandafter\ifx\csname urlstyle\endcsname\relax
  \providecommand{\doi}[1]{doi: #1}\else
  \providecommand{\doi}{doi: \begingroup \urlstyle{rm}\Url}\fi

\bibitem[Bear(2016)]{thesis}
Helen~L Bear.
\newblock \emph{Decoding visemes: improving machine lip-reading. PhD thesis}.
\newblock University of East Anglia, 2016.

\bibitem[Bear and Harvey(2016)]{bear2016decoding}
Helen~L Bear and Richard Harvey.
\newblock Decoding visemes: improving machine lip-reading.
\newblock In \emph{41st International Conference Acoustics, Speech and Signal
  Processing (ICASSP)}, 2016.

\bibitem[Bear et~al.(2014{\natexlab{a}})Bear, Harvey, Theobald, and
  Lan]{bear2014resolution}
Helen~L Bear, Richard Harvey, Barry-John Theobald, and Yuxuan Lan.
\newblock Resolution limits on visual speech recognition.
\newblock In \emph{Image Processing (ICIP), IEEE International Conference on},
  pages 1371--1375. IEEE, 2014{\natexlab{a}}.

\bibitem[Bear et~al.(2014{\natexlab{b}})Bear, Harvey, Theobald, and
  Lan]{bear2014phoneme}
Helen~L Bear, Richard~W Harvey, Barry-John Theobald, and Yuxuan Lan.
\newblock Which phoneme-to-viseme maps best improve visual-only computer
  lip-reading?
\newblock In \emph{Advances in Visual Computing}, pages 230--239. Springer,
  2014{\natexlab{b}}.
\newblock \doi{10.1007/978-3-319-14364-4_22}.

\bibitem[Bear et~al.(2014{\natexlab{c}})Bear, Owen, Harvey, and
  Theobald]{bear2014obs}
Helen~L Bear, Gari Owen, Richard Harvey, and Barry-John Theobald.
\newblock Some observations on computer lip-reading: moving from the dream to
  the reality.
\newblock In \emph{SPIE Security and Defence}, pages 92530G--92530G.
  International Society for Optics and Photonics, 2014{\natexlab{c}}.
\newblock \doi{10.1117/12.2067464}.

\bibitem[Bear et~al.(2015)Bear, Cox, and Harvey]{bear2015speakerindep}
Helen~L Bear, Stephen~J Cox, and Richard Harvey.
\newblock Speaker independent machine lip reading with speaker dependent viseme
  classifiers.
\newblock In \emph{1st Joint International Conference on Facial Analysis,
  Animation and Audio-Visual Speech Processing (FAAVSP)}, pages 115--120. ISCA,
  2015.

\bibitem[{Cambridge University, UK}(1997)]{beep}
{Cambridge University, UK}.
\newblock The {BEEP} {Pronunciation} {Dictionary} - {British} {English}.
\newblock \url{ftp://svr-ftp.eng.cam.ac.uk/comp.speech/dictionaries/}, 1997.
\newblock Accessed: January 2013.

\bibitem[Cappelletta and Harte(2012)]{cappelletta2012phoneme}
Luca Cappelletta and Naomi Harte.
\newblock Phoneme-to-viseme mapping for visual speech recognition.
\newblock In \emph{International Conference on Pattern Recognition Applications
  and Methods (ICPRAM)}, pages 322--329, 2012.

\bibitem[Cetingul et~al.(2006)Cetingul, Yemez, Erzin, and Tekalp]{1703580}
H.~E. Cetingul, Y.~Yemez, E.~Erzin, and A.~M. Tekalp.
\newblock Discriminative analysis of lip motion features for speaker
  identification and speech-reading.
\newblock \emph{IEEE Transactions on Image Processing}, 15\penalty0
  (10):\penalty0 2879--2891, Oct 2006.
\newblock ISSN 1057-7149.
\newblock \doi{10.1109/TIP.2006.877528}.

\bibitem[Chung and Zisserman(2016)]{chungaccv}
J~S Chung and Andrew Zisserman.
\newblock Lip reading in the wild.
\newblock In \emph{Asian Conference on Computer Vision}, 2016.

\bibitem[Cox et~al.(2008)Cox, Harvey, Lan, Newman, and
  Theobald]{cox2008challenge}
Stephen Cox, Richard Harvey, Yuxuan Lan, Jacob Newman, and Barry Theobald.
\newblock The challenge of multispeaker lip-reading.
\newblock In \emph{Proceedings of the International Conference on
  Auditory-Visual Speech Processing (AVSP)}, pages 179--184, 2008.

\bibitem[Efron and Gong(1983)]{efron1983leisurely}
Bradley Efron and Gail Gong.
\newblock A leisurely look at the bootstrap, the jackknife, and
  cross-validation.
\newblock \emph{The American Statistician}, 37:\penalty0 36--48, 1983.

\bibitem[Hazen et~al.(2004)Hazen, Saenko, La, and Glass]{Hazen1027972}
Timothy~J. Hazen, Kate Saenko, Chia-Hao La, and James~R. Glass.
\newblock A segment-based audio-visual speech recognizer: Data collection,
  development, and initial experiments.
\newblock In \emph{Proceedings of the 6th International Conference on
  Multimodal Interfaces}, ICMI `04, pages 235--242, New York, NY, USA, 2004.
  ACM.
\newblock ISBN 1-58113-995-0.
\newblock \doi{10.1145/1027933.1027972}.
\newblock URL \url{http://doi.acm.org/10.1145/1027933.1027972}.

\bibitem[Kleinschmidt and Jaeger(2015)]{kleinschmidt2015robust}
Dave~F Kleinschmidt and T~Florian Jaeger.
\newblock Robust speech perception: Recognize the familiar, generalize to the
  similar, and adapt to the novel.
\newblock \emph{Psychological review}, 122\penalty0 (2):\penalty0 148, 2015.

\bibitem[Lan et~al.(2012)Lan, Harvey, and Barry-John]{lan2012insights}
Yuxuan Lan, Richard Harvey, and Theobald Barry-John.
\newblock Insights into machine lip reading.
\newblock In \emph{38th International Conference Acoustics, Speech and Signal
  Processing (ICASSP)}, 2012.

\bibitem[Lesner and Kricos(1981)]{visualvowelpercept}
S.A. Lesner and P.B Kricos.
\newblock Visual vowel and diphthong perception across speakers.
\newblock \emph{Journal of the Academy of Rehabilitative Audiology},
  14:\penalty0 252--258, 1981.

\bibitem[Lucey and Potamianos(2006)]{4064511}
P.~Lucey and G.~Potamianos.
\newblock Lipreading using profile versus frontal views.
\newblock In \emph{2006 IEEE Workshop on Multimedia Signal Processing}, pages
  24--28, Oct 2006.
\newblock \doi{10.1109/MMSP.2006.285261}.

\bibitem[Lucey et~al.(2009)Lucey, Potamianos, and Sridharan]{lucey2009visual}
Patrick Lucey, Gerasimos Potamianos, and Sridha Sridharan.
\newblock Visual speech recognition across multiple views.
\newblock \emph{Visual Speech Reognition: Lip Segmentation and Mapping}, 2009.

\bibitem[Luettin et~al.(1996)Luettin, Thacker, and Beet]{607030}
J.~Luettin, N.~A. Thacker, and S.~W. Beet.
\newblock Speaker identification by lipreading.
\newblock In \emph{Spoken Language, 1996. ICSLP 96. Proceedings., Fourth
  International Conference on}, volume~1, pages 62--65 vol.1, Oct 1996.
\newblock \doi{10.1109/ICSLP.1996.607030}.

\bibitem[Matthews and Baker(2004)]{Matthews_Baker_2004}
Iain Matthews and Simon Baker.
\newblock Active appearance models revisited.
\newblock \emph{International Journal of Computer Vision}, 60:\penalty0
  135--164, 2004.

\bibitem[Taylor et~al.(2012)Taylor, Mahler, Theobald, and
  Matthews]{taylor2012dynamic}
Sarah~L Taylor, Moshe Mahler, Barry-John Theobald, and Iain Matthews.
\newblock Dynamic units of visual speech.
\newblock In \emph{Proceedings of the 11th ACM SIGGRAPH/Eurographics conference
  on Computer Animation}, pages 275--284. Eurographics Association, 2012.

\bibitem[Thangthai et~al.(2015)Thangthai, Harvey, Cox, and
  Theobald]{thangthai2015improving}
Kwanchiva Thangthai, Richard~W Harvey, Stephen~J Cox, and Barry-John Theobald.
\newblock Improving lip-reading performance for robust audiovisual speech
  recognition using dnns.
\newblock In \emph{AVSP}, pages 127--131, 2015.

\bibitem[Wong et~al.(2011)Wong, ChÕng, Seng, Ang, Chin, Chew, and
  Lim]{Wong20111503}
Yee~Wan Wong, Sue~Inn ChÕng, Kah~Phooi Seng, Li-Minn Ang, Siew~Wen Chin,
  Wei~Jen Chew, and King~Hann Lim.
\newblock A new multi-purpose audio-visual unmc-vier database with multiple
  variabilities.
\newblock \emph{Pattern Recognition Letters}, 32\penalty0 (13):\penalty0 1503
  -- 1510, 2011.
\newblock ISSN 0167-8655.
\newblock \doi{http://dx.doi.org/10.1016/j.patrec.2011.06.011}.

\bibitem[Young et~al.(2006)Young, Evermann, Gales, Kershaw, Moore, Odell,
  Ollason, Povey, Valtchev, and Woodland]{young2006htk}
Steve~J Young, Gunnar Evermann, MJF Gales, D~Kershaw, G~Moore, JJ~Odell,
  DG~Ollason, D~Povey, V~Valtchev, and PC~Woodland.
\newblock The {HTK} book version 3.4, 2006.

\end{thebibliography}
\end{document}